\documentclass[10pt,twocolumn,letterpaper]{article}

\usepackage[pagenumbers]{cvpr} 

\usepackage{graphicx}
\usepackage{booktabs}
\usepackage{placeins}

\definecolor{SBFMAccent}{HTML}{B5432A}
\definecolor{SBFMRow}{HTML}{FFF1E8}
\makeatletter
\newcommand{\SBFMTableRow}{%
  \noalign{\begingroup\color{SBFMRow}%
    \hrule height\dimexpr\ht\@arstrutbox+\dp\@arstrutbox\relax
    \kern-\dimexpr\ht\@arstrutbox+\dp\@arstrutbox\relax
  \endgroup}%
}
\makeatother

\usepackage{microtype}

\DeclareMathOperator{\wrap}{wrap}
\DeclareMathOperator{\sg}{sg}
\newcommand{\proj}{\mathcal{P}}

\definecolor{cvprblue}{rgb}{0.21,0.49,0.74}
\usepackage[pagebackref,breaklinks,colorlinks,allcolors=cvprblue]{hyperref}

\usepackage{cuted}

\def\confName{CVPR}
\def\confYear{2027}

\title{S-BFM: A Behavioral Foundation Model for Humanoid Control\\ from Sparse Spatiotemporal Goals}

\author{
Kun Yang$^{1,4,*}$ \quad
Xuanyu Chen$^{2,4,*}$ \quad
Jing Tan$^{1}$ \quad
Yanjing Liu$^{1}$ \\
Yubo Zhao$^{3}$ \quad
Renjing Xu$^{1,\dagger}$ \quad
Addison Lin Wang$^{2,\ddagger}$ \\
$^{1}$The Hong Kong University of Science and Technology (Guangzhou) \\
$^{2}$Nanyang Technological University \\
$^{3}$Shenzhen MSU-BIT University \\
$^{4}$AGIBOT \\
{\tt\small
kyang556@connect.hkust-gz.edu.cn,
XUANYU005@e.ntu.edu.sg
} \\
{\tt\small
colinjing1@gmail.com,
yliu449@connect.hkust-gz.edu.cn,
1120230573@smbu.edu.cn
} \\
{\tt\small
renjingxu@hkust-gz.edu.cn,
linwang@ntu.edu.sg
} \\
$^{*}$Equal contribution. \\
$^{\dagger}$First corresponding author. \quad
$^{\ddagger}$Second corresponding author.
}

\begin{document}
\raggedbottom
\maketitle

\begin{strip}
\vspace{-40pt}
    \centering
    \includegraphics[width=\textwidth]{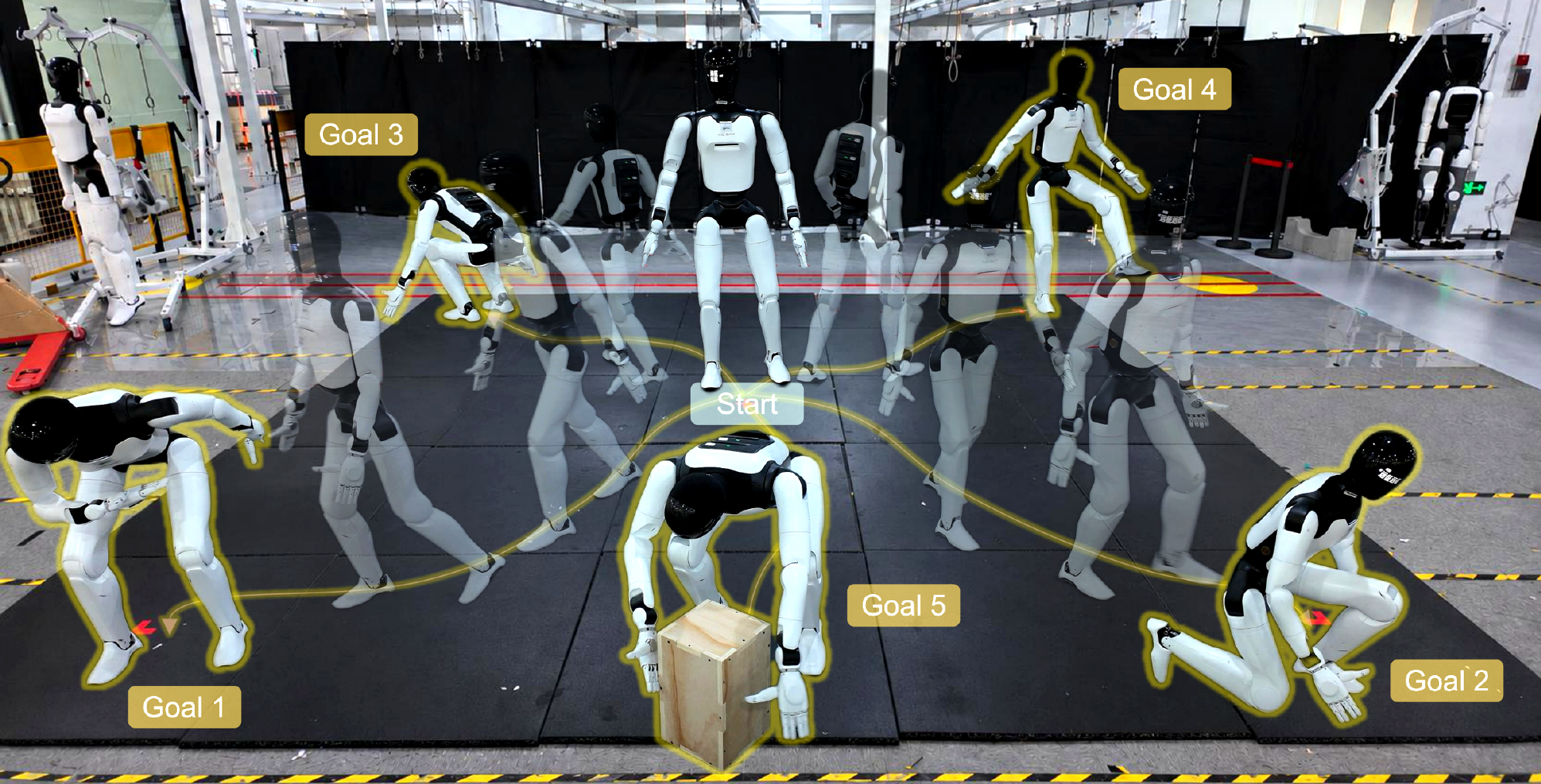}
    \vspace{-15pt}
    \captionsetup{hypcap=false}\captionof{figure}{\textbf{S-BFM on a real humanoid.}
    Given sparse spatiotemporal goals, the robot autonomously transitions from a shared initial pose to five target postures. Translucent silhouettes show intermediate poses, while gold outlines mark the target postures and positions.}
    \label{fig:sbfm_teaser}
    \vspace{-8pt}
\end{strip}

\begin{abstract}
General-purpose whole-body trackers can execute diverse humanoid motions, but they require dense reference trajectories.
However, tasks usually specify only where, in which posture, and when the robot should arrive, leaving the motion in between to a separate planner.
Inspired by human motor coordination,\textbf{ our key idea} is that {a low-level controller can organize the intervening motion and its pace by itself from sparse goals with arrival times}.
In this paper, we propose \textbf{S-BFM}, a novel behavioral foundation model superior in its autonomy: it generates whole-body transitions from target postures, poses, and arrival times alone.
Specifically, we first introduce \textbf{Spatial Grounding}, which adds world-frame position and heading to a forward-backward behavior representation through spatial reconstruction.
Then, \textbf{Time-Conditioned Distillation} is designed to learn transitions and pace from real motion through closed-loop imitation of a dense-reference teacher.
Lastly, \textbf{Generative Fine-Tuning} is introduced to broaden the covered distances and arrival times with a distilled motion generator.
In simulation, S-BFM lowers global joint-position error by over \textbf{32\%} relative to the best baseline under fixed and mixed horizons, and raises fixed-goal reach from 43.5\% to \textbf{74.6\%} while cutting terminal position error from 1.67 to \textbf{0.26}\,m.
We further deploy S-BFM on the full-size AGIBOT A3 humanoid.
Sparse spatiotemporal goals thus suffice for natural, well-timed whole-body motion.
\end{abstract}    
\vspace{-15pt}
\section{Introduction}
\label{sec:intro}

Humanoid control is moving from task-specific skill policies toward general-purpose whole-body control.
Large motion datasets and physics simulation have enabled universal trackers that turn diverse motion references into executable joint commands.
These trackers connect motion generators and multimodal interfaces to the robot.
TextOp continuously generates short motion segments for a tracking policy to execute~\cite{xie2026textop}, SONIC links input modalities to control through a unified motion interface~\cite{luo2026sonic}, and SuperMimic trains whole-body trackers efficiently with a shared differentiable world model~\cite{liao2026supermimic}.
These trackers can execute diverse humanoid motions, but they require dense reference trajectories.

However, task planning and interactive control require robots to organize motion around task goals and adapt as those goals change.
With dense-reference interfaces, this responsibility falls to a separate planner or generator.
It must supply the intervening motion and update future references when the task changes while ensuring a smooth transition from the motion already underway~\cite{luo2026sonic,xie2026textop}.
Human movement instead combines goal-directed decisions with automatic coordination.
For example, when walking toward a table, the cerebral cortex helps select the goal and route.
Brainstem--spinal circuits support gait and posture while the cerebellum uses sensory feedback to coordinate movement.
Together, these mechanisms allow people to focus on their destination and surroundings without consciously controlling every step~\cite{clark2015automaticity}.
Inspired by this coordination, we ask a novel question: \emph{Can a low-level controller accept sparse goals with arrival times and organize the intervening motion from its current physical state?}

Prior work provides foundations:
BFM-Zero supports goal reaching and motion tracking via a shared behavior space~\cite{li2026bfmzero}, and BeJEPA models whole-body posture together with world position and heading~\cite{yan2026bejepa}.
MaskedMimic and KINO produce physics-based motion from partial constraints or keyframes~\cite{tessler2024maskedmimic,chen2026kino}, and BFMTrack tracks sparse timed goals by latent optimization~\cite{rupf2026bfmtrack}.
Yet no existing method delivers all three sparse-goal requirements at once: accurate control of the world-frame pose, natural transitions between goals, and timing that adapts to the requested arrival.
A target state alone does not specify how the intervening motion should unfold, and imitating the pace of a demonstration does not adapt it to new distances and durations.

Inspired by human motor coordination, our key idea is that \textbf{a low-level controller can organize the intervening motion and its pace by itself from sparse goals with arrival times}.
In this paper, we propose \textbf{S-BFM} (Sparse Behavioral Foundation Model), which is superior in its autonomy: it generates whole-body transitions from target postures, poses, and arrival times alone (\cref{fig:sbfm_teaser}).
S-BFM is trained in three stages within a reusable forward-backward behavior space~\cite{touati2021fb,li2026bfmzero,han2026ibfm,wang2026pbfm}.
Each stage targets one requirement.
Specifically, we first propose \textbf{Spatial Grounding} (\cref{sec:spatial}), which adds world-frame position and heading to the behavior representation and uses spatial reconstruction to make the encoder retain them.
Then, \textbf{Time-Conditioned Distillation} (\cref{sec:temporal}) is designed to learn transitions and pace from real motion: with the encoder and actor frozen, a controller conditioned on the remaining time imitates a dense-reference teacher in closed loop.
As real demonstrations cover only a limited range of distances and durations, \textbf{Generative Fine-Tuning} (\cref{sec:generative}) is lastly introduced to broaden this range with a lightweight motion generator distilled from Kimodo~\cite{rempe2026kimodo}.
At deployment, S-BFM acts from robot observations, sparse goals, and the remaining time, without any intermediate reference.

We evaluate S-BFM in simulation on a 29-DoF Unitree G1 against SONIC, UFO, BeJEPA, and PLAT, and on the AGIBOT A3 humanoid.
Under both fixed and mixed command horizons, S-BFM lowers global joint-position error by over \textbf{32\%} relative to the best baseline.
For fixed goals, it raises the reach rate from 43.5\% to \textbf{74.6\%} over BeJEPA, increases the rate of holding the goal for five seconds from 7.4\% to \textbf{35.2\%}, and cuts terminal position error from 1.67 to \textbf{0.26}\,m.
At 3 to 4\,m under lateral and downward pushes, it still reaches 68.0\% of goals, against 23.6\% for BeJEPA.
On the real robot, S-BFM moves from a shared start pose to five distinct goal postures (\cref{fig:sbfm_teaser}).
Sparse spatiotemporal goals thus suffice for natural, well-timed whole-body motion.

\noindent In summary, our main contributions are threefold:

\noindent \textbf{1) }We formulate sparse spatiotemporal goal control, in which a low-level policy, much like the cerebellum, generates natural whole-body transitions that reach target postures at prescribed world poses and times.

\noindent \textbf{2) }We propose S-BFM, trained through Spatial Grounding, Time-Conditioned Distillation, and Generative Fine-Tuning, which combines spatial goal reaching with the transitions and pace of real motion.

\noindent \textbf{3)} In simulation, S-BFM outperforms all evaluated baselines in sparse tracking accuracy, spatial goal reaching, and goal retention, with over 32\% lower global joint-position error than the best baseline, and it runs on the full-size AGIBOT A3 humanoid.

\vspace{-5pt}
\section{Related Work}
\label{sec:related_work}

\noindent\textbf{Dense-reference whole-body control.}
Physics-based motion imitation learns whole-body skills by tracking reference trajectories in simulation~\cite{peng2018deepmimic,luo2023phc,chen2025gmtgeneralmotiontracking,tan2026slmp}.
H2O, TWIST, and ExBody series enable versatile humanoid motion tracking and teleoperation~\cite{he2024learning,he2024omnih2o,ze2025twistteleoperatedwholebodyimitation,ze2025twist2,cheng2024exbody,ji2024exbody2}.
SONIC scales tracking with model capacity, data, and compute~\cite{luo2026sonic}, SuperMimic and BlinkMimic train trackers with differentiable world models~\cite{liao2026supermimic,chen2026blinkmimictraininghumanoidmotion}, and other work extends whole-body control to rough terrain and highly dynamic skills~\cite{li2026pgmt,ma2026extremergmt,chen2026physicsguidedbiomechanicalgaitadaptation,hu2026maskillblenderdecentralizedwholebodycoordination,tan2026farm,he2025asap,harithas2026braceadaptingwholebodyreferences}.
These trackers are general, but \textit{they execute a reference rather than create one}: given only sparse goals, they still need another module to supply the motion in between. In contrast,
S-BFM needs no such module at deployment.

\noindent\textbf{Sparse and time-conditioned motion control.}
Sparse control specifies only desired states and leaves the intermediate motion to the controller~\cite{zargarbashi2024robotkeyframinglearninglocomotionhighlevel,huang2025adaptablehumanoidcontroladaptive}.
Goal-conditioned value functions and temporal difference models support goal- and horizon-conditioned control~\cite{schaul2015uvfa,pong2018tdm}.
Forward-backward representations enable zero-shot goal reaching and tracking, as demonstrated by BFM-Zero~\cite{touati2021fb,touati2023zeroshot,li2026bfmzero}.
BeJEPA adds world position and heading to body representation~\cite{yan2026bejepa}.
MaskedMimic inpaints physics-based motion from partial constraints~\cite{tessler2024maskedmimic}, KINO links vision-language planning to control through whole-body keyframes~\cite{chen2026kino}, PLAT learns privileged latent transitions for sparse timed tracking~\cite{wang2026plat}, and BFMTrack optimizes latent behavior sequences~\cite{rupf2026bfmtrack}.
Unfortunately, \textit{none of these methods delivers accurate world-frame pose control, natural transitions, and adaptive timing at once}.
S-BFM addresses all three within one forward-backward behavior space, with \textit{one stage per requirement}.

\noindent\textbf{Motion in-betweening and controllable generation.}
Motion in-betweening generates transitions between constrained states.
Robust Motion In-Betweening conditions on time-to-arrival embeddings~\cite{harvey2020inbetweening}, diffusion models such as MDM and CondMDI generate motion under keyframe constraints~\cite{tevet2023mdm,cohan2024condmdi}, Kimodo supports text and kinematic constraints~\cite{rempe2026kimodo,yang2026uniquenft}, and ARDY handles streaming prompts and sparse constraints~\cite{zhao2026ardy}.
To connect generation with execution, TextOp pairs a generator with a tracker~\cite{xie2026textop}, InterPrior learns generative control for physical interaction~\cite{xu2026interprior,ye2026predactor}, and GRAIL generates interaction demonstrations with video world models~\cite{xie2026grail}.
However, \textit{a kinematically valid motion does not guarantee that the target pose and timing are kept during closed-loop physical execution}.
S-BFM therefore uses a generator only during training, to broaden the distances and durations its controller learns from.
\section{Method}
\label{sec:method}

\begin{figure*}[t]
    \centering
    \includegraphics[width=\textwidth]{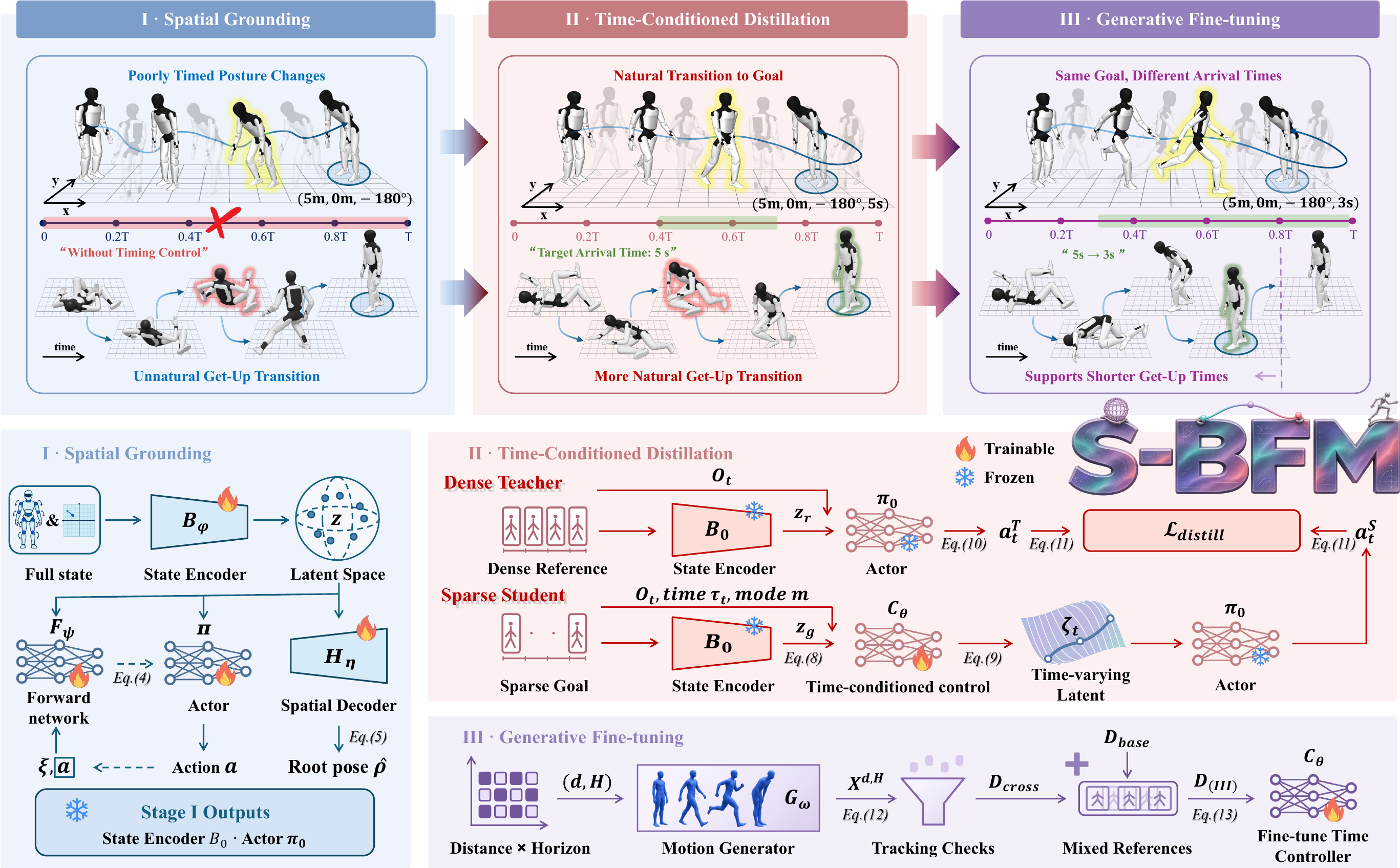}
    \caption{\textbf{Overview of our S-BFM framework.}
    Stage~I learns a behavior representation grounded in world-frame position and heading.
    Stage~II distills reference-guided transitions conditioned on the remaining time.
    Stage~III broadens the range of distances and horizons through generative fine-tuning.}
    \label{fig:sbfm_architecture}
\end{figure*}

Our goal is a single controller that receives a target posture, position, heading, and arrival time, and produces the whole-body motion that gets there on time.
We first define this problem (\cref{sec:sbfm_problem}) and then describe how S-BFM learns it in three stages (\cref{fig:sbfm_architecture}).

\vspace{-1pt}
\subsection{Problem Formulation}
\label{sec:sbfm_problem}

Each goal $g$ specifies a target body state at a planar position and heading, together with an arrival time $T$.
The controller must generate the intervening motion and reach $g$ at time $T$.
The full state $x_t$ contains the whole-body configuration and motion together with the root pose, and the control observation $o_t$ contains proprioception, the estimated root pose, and a short history of observations and actions.

\noindent\textbf{Reference frame.}
Goals are placed in the world, so the robot and all goals share a fixed frame $\mathcal{W}$ defined at the start of execution.
Its origin is the initial planar root position $p_{0,xy}$, and its reference direction is the initial heading $\psi_0$:
\begin{equation}
\begin{aligned}
p_{t,xy}^{W} &= R(-\psi_0)(p_{t,xy}-p_{0,xy}),\\
\psi_t^{W} &= \wrap(\psi_t-\psi_0),
\end{aligned}
\label{eq:sbfm_frame}
\end{equation}
where $R$ is a planar rotation and $\wrap$ maps angles to $[-\pi,\pi)$.

\noindent\textbf{Timing.}
The controller is told how much time remains,
\begin{equation}
\Delta_t = \max(T-t,0),
\label{eq:sbfm_remaining}
\end{equation}
which decreases while the goal stays fixed; a new command replaces both $g$ and $T$.
A binary mode $m$ states whether the robot should hold the goal or continue after arrival.
At deployment, the controller maps $(o_t, g, \Delta_t, m)$ to joint targets for a low-level PD controller.
Dense reference motions are available only during training.

\subsection{Overview}
\label{sec:sbfm_overview}

Reaching a sparse goal on time requires the controller to know where the goal is and how to move toward it at the right pace.
S-BFM learns this in three stages (\cref{fig:sbfm_architecture}).
Stage~I (\cref{sec:spatial}) builds a forward-backward behavior representation that retains world-frame position and heading.
Stage~II (\cref{sec:temporal}) freezes this representation and trains a time-conditioned controller to master the transitions and pace of real motion.
Stage~III (\cref{sec:generative}) fine-tunes the controller on generated motions that cover distances and time horizons missing from real data.

The stages are trained in sequence rather than jointly so that later stages do not overwrite what earlier ones learned.
After Stage~I, the encoder and actor are frozen, and Stages~II and~III train only a controller on top of them.
At inference, the encoder maps the goal to a goal code, the controller combines this code with $o_t$, $\Delta_t$, and $m$, and the actor turns the resulting behavior code into joint actions.
The spatial readout, the dense teacher, and the motion generator are used only during training.

\subsection{Stage I: Spatial Grounding}
\label{sec:spatial}

We build on the forward-backward (FB) behavior model of BFM-Zero~\cite{li2026bfmzero} for whole-body behaviors in a shared latent space.
Its behavior codes are not trained to encode where the robot is in the world, so we add world-frame root pose to both the representation and training signal.

\noindent\textbf{Root-pose features.}
In the frame of \cref{eq:sbfm_frame}, root position and heading are encoded as
\begin{equation}
\rho(x_t)=\left[
\frac{p_{t,x}^{W}}{s_p},\frac{p_{t,y}^{W}}{s_p},
\sin\psi_t^{W},\cos\psi_t^{W}\right],
\label{eq:sbfm_root_feature}
\end{equation}
where $s_p$ scales position and the sine-cosine pair avoids the discontinuity at $\pm\pi$.
These features are appended to the encoder input and to the control observation and its history.

\noindent\textbf{Behavior learning.}
The backward encoder $B_\phi$ maps a full state to a behavior embedding, which is projected to a behavior code by $\proj(v)=\sqrt{d_z}\,v/\lVert v\rVert_2$, with $d_z$ the latent dimension.
The forward network $F_\psi$ predicts long-term state visitation under a behavior code $z$:
\begin{equation}
\begin{aligned}
M_z(\xi,a,x') &= F_\psi(\xi,a,z)^\top B_\phi(x'),\\
Q_z(\xi,a) &= F_\psi(\xi,a,z)^\top z,
\end{aligned}
\label{eq:sbfm_fb}
\end{equation}
where $\xi$ is the training observation.
We retain the FB temporal-difference objective $\mathcal{L}_{\mathrm{FB}}$ on simulator transitions, together with motion prior and physical regularization of BFM-Zero.
The property of this model is that encoding a target state, $z=\proj(B_\phi(g))$, yields a code whose policy moves the robot toward that state; Stage~I extends this property from body posture to world-frame position and heading.

\noindent\textbf{Spatial supervision.}
To make the embedding retain root pose, a readout $H_\eta$ reconstructs the normalized root-pose features from it:
\begin{equation}
\begin{gathered}
\mathcal{L}_{\mathrm{sp}}
=\mathbb{E}_{x\sim\mathcal{D}_{\mathrm{sp}}}
\left[\tfrac14\left\lVert W^{1/2}e_\rho(x)\right\rVert_2^2\right],\\
e_\rho(x)
=H_\eta\!\left(\proj(B_\phi(x))\right)-\widetilde\rho(x),
\end{gathered}
\label{eq:sbfm_spatial_loss}
\end{equation}
where $\widetilde\rho$ is the normalized target and $W$ balances position against heading.
This loss updates both the readout and backward encoder.
The data $\mathcal{D}_{\mathrm{sp}}$ mix robot-visited states and reference goals in equal proportion, with goal positions and headings randomized near the robot.
Each reference segment is moved by a single planar rigid transformation, keeping its internal geometry and kinematics. Also, these samples are used only for $\mathcal{L}_{\mathrm{sp}}$, not for FB learning.
The full objective is as follows:
\begin{equation}
\mathcal{L}_{\mathrm{repr}}
=\mathcal{L}_{\mathrm{FB}}
+\lambda_{\mathrm{orth}}\mathcal{L}_{\mathrm{orth}}
+\lambda_{\mathrm{sp}}\mathcal{L}_{\mathrm{sp}},
\label{eq:sbfm_representation_loss}
\end{equation}
where $\mathcal{L}_{\mathrm{orth}}$ is the orthogonality regularizer of FB.
Stage~I produces the state encoder $B_0$ and the actor $\pi_0$, which stay frozen afterwards.

\subsection{Stage II: Time-Conditioned Distillation}
\label{sec:temporal}

After Stage~I, the policy can reach a goal, but the goal state alone does not determine the path or the speed.
Stage~II learns both from real motion by distilling a teacher that follows a dense reference into a student that sees only the sparse goal and the remaining time.

\noindent\textbf{Training tasks.}
From a reference motion, we sample a start time $u_s$ and a goal time $u_g$.
The state at $u_g$ becomes the sparse goal $g$, and the segment duration sets the arrival time,
\begin{equation}
T=t_s+(u_g-u_s),
\label{eq:sbfm_reference_timing}
\end{equation}
where $t_s$ denotes execution start.
The reference is replayed at native speed and supplies the teacher's intermediate targets.

\noindent\textbf{Student.}
We freeze $B_0$, $\pi_0$, and the observation normalizer and train only a controller $C_\theta$.
The goal is encoded once as
\begin{equation}
z_g=\proj(B_0(g)),
\label{eq:sbfm_goal_code}
\end{equation}
and the remaining time is normalized as $\tau_t=\Delta_t/s_\tau$.
At every step, the controller produces a behavior code, which the frozen actor executes:
\begin{equation}
\zeta_t=\proj\!\left(C_\theta(o_t,z_g,\tau_t,m)\right),
\qquad
a_t^S=\pi_0(o_t,\zeta_t).
\label{eq:sbfm_control_code}
\end{equation}
The goal code $z_g$ is constant between keyframes, while $\zeta_t$ adapts to the current observation and the time left.

\noindent\textbf{Teacher.}
The teacher uses the same observation, replacing the sparse goal with a time-aligned reference state $r_t$:
\begin{equation}
a_t^T=\pi_0\!\left(o_t,\proj(B_0(r_t))\right).
\label{eq:sbfm_teacher_action}
\end{equation}
Since the teacher sees where the reference is at each moment while the student sees only the endpoint and the time left, matching the teacher's actions teaches the student both the transition and its pace.

\noindent\textbf{Closed-loop distillation.}
As in DAgger~\cite{ross2011dagger}, the student executes its own actions and the teacher labels the states the student visits, so the student also learns to recover from its own drift.
Paired inputs and teacher actions are stored in a buffer $\mathcal{B}_{\mathrm{IL}}$, and the controller minimizes
\begin{equation}
\mathcal{L}_{\mathrm{distill}}
=\mathbb{E}_{\mathcal{B}_{\mathrm{IL}}}
\left[\frac{\left\lVert a_t^S-\sg(a_t^T)\right\rVert_2^2}{n_a}\right],
\label{eq:sbfm_distillation_loss}
\end{equation}
where $n_a$ is the action dimension and $\sg$ is the stop-gradient.
Gradients flow through the frozen actor into $C_\theta$, which is initialized to output $z_g$ so that training starts from goal-reaching behavior of Stage~I.
For hold commands, the goal stays active after $T$ with zero remaining time, and the teacher continues to supervise the robot as it holds the pose.

\subsection{Stage III: Generative Fine-Tuning}
\label{sec:generative}

Real motion pairs each transition with one distance and one time horizon, whereas at deployment the same pair of postures may be requested at other distances and horizons.
Stage~III fills these gaps with generated motion and fine-tunes the controller on it.

\noindent\textbf{Distance-horizon expansion.}
We distill a pretrained Kimodo model~\cite{rempe2026kimodo} into a compact conditional motion generator $\mathcal{G}_\omega$ that accepts a text description together with spatial and temporal constraints.
For an initial state $x_s$ and a goal $g$, we vary the travel distance $d$ and horizon $H$ and generate
\begin{equation}
\begin{aligned}
\bar X^{d,H}
&=\mathcal{G}_\omega(\ell_{d,H},\mathcal{K}_{d,H},\epsilon),\\
\mathcal{K}_{d,H}
&=\mathcal{K}(x_s,\mathcal{T}_d(g),H),
\end{aligned}
\label{eq:sbfm_generation}
\end{equation}
where $\mathcal{T}_d$ moves the target to distance $d$ without changing its posture or heading, $\mathcal{K}$ attaches the endpoint pose and position constraints to their times, and $\epsilon$ is sampling noise.
The text $\ell_{d,H}$ comes from reusable templates that describe the motion and its pace, while the numerical constraints fix distance and horizon.
For the same endpoint postures, different $(d,H)$ thus produce different intermediate postures and velocities, which supervise the controller across remaining-time conditions.

\begin{table*}[t]
\centering
\caption{\textbf{Sparse timed keyframe tracking under fixed and mixed command horizons.}}
\label{tab:q1_sparse}
\small
\setlength{\tabcolsep}{4pt}
\begin{tabular*}{\textwidth}{@{\extracolsep{\fill}}lrrrrrrrrrr@{}}
\toprule
& \multicolumn{5}{c}{Fixed horizons: $0.1$--$3.5$\,s} & \multicolumn{5}{c}{Mix: $0.02$--$2.5$\,s} \\
\cmidrule(lr){2-6}\cmidrule(lr){7-11}
Method & \shortstack{Local\\(cm)$\downarrow$} & \shortstack{Global\\(cm)$\downarrow$} & \shortstack{XY\\(m)$\downarrow$} & \shortstack{Yaw\\($^\circ$)$\downarrow$} & \shortstack{AR\\(\%)$\uparrow$} & \shortstack{Local\\(cm)$\downarrow$} & \shortstack{Global\\(cm)$\downarrow$} & \shortstack{XY\\(m)$\downarrow$} & \shortstack{Yaw\\($^\circ$)$\downarrow$} & \shortstack{AR\\(\%)$\uparrow$} \\
\midrule
SONIC~\cite{luo2026sonic} & 13.00 & 158.86 & 1.558 & 24.72 & 35.73 & 14.29 & 173.67 & 1.704 & 27.15 & 33.99 \\
UFO~\cite{ufo2026} & 14.36 & 187.87 & 1.859 & 53.08 & 32.77 & 15.18 & 201.71 & 1.996 & 56.83 & 31.93 \\
BeJEPA~\cite{yan2026bejepa} & 10.20 & 59.35 & 0.572 & 20.90 & 77.89 & 10.57 & 57.93 & 0.555 & 21.73 & 77.65 \\
PLAT~\cite{wang2026plat} & 16.61 & 70.45 & 0.639 & 38.38 & 71.33 & 14.29 & 55.49 & 0.498 & 30.60 & 79.26 \\
\SBFMTableRow
\textcolor{SBFMAccent}{\textbf{S-BFM (ours)}} & \textbf{8.87} & \textbf{38.35} & \textbf{0.362} & \textbf{16.80} & \textbf{85.47} & \textbf{8.88} & \textbf{37.60} & \textbf{0.354} & \textbf{16.38} & \textbf{86.04} \\
\bottomrule
\end{tabular*}
\end{table*}

\begin{figure*}[t]
    \centering
    \includegraphics[width=\textwidth]{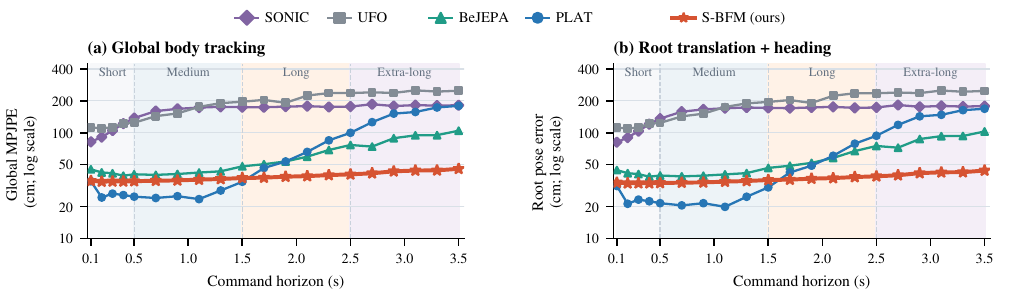}
    \caption{\textbf{Tracking error versus command horizon.} Both panels use logarithmic error scales. Root-pose error is defined in Eq.~\ref{eq:q1_pose_error}.}
    \label{fig:q1_horizon}
\end{figure*}

\noindent\textbf{Fine-tuning.}
Generated motions are kept only if they pass boundary-consistency processing and physical tracking checks.
Accepted motions enter a reference buffer that is sampled together with the real references,
\begin{equation}
\mathcal{D}_{\mathrm{III}}
=(1-\alpha)\mathcal{D}_{\mathrm{base}}
+\alpha\mathcal{D}_{\mathrm{cross}},
\label{eq:sbfm_data_mixture}
\end{equation}
where $\mathcal{D}_{\mathrm{cross}}$ holds the references expanded across distances and horizons and $\alpha$ is the mixing ratio.
Starting from the Stage~II controller, with the encoder and actor still frozen, we fine-tune with \cref{eq:sbfm_distillation_loss} on teacher targets from this mixture.
Generated references with new timing use a 0.6\,s teacher preview. All other references keep the one-step (0.02\,s) target.
The generator is used only to create training references, so deployment keeps the interface of \cref{sec:sbfm_problem}.

\section{Experiments}
\label{sec:experiments}
\begingroup
\setlength{\parskip}{0pt}
\flushbottom We evaluate S-BFM through four questions:
\textbf{Q1:} Can S-BFM jointly track posture and world-space position from sparse timed keyframes?
\textbf{Q2:} Can S-BFM autonomously reach and maintain spatial goals across distances and perturbations without intermediate reference trajectories?
\textbf{Q3:} What control capabilities do spatial grounding, time-conditioned distillation, and generative fine-tuning each contribute?
\textbf{Q4:} How does S-BFM perform on physical robots?

\subsection{Experimental Setup}
\label{sec:experimental_setup}
\noindent\textbf{Environment and data.}\enspace
Simulation uses a 29-DoF Unitree G1 in MuJoCo at 50\,Hz with a 2\,ms physics timestep.
Q1--Q4 use evaluation tasks constructed from LAFAN and BONES-SEED.
Within each experiment, methods share test data, goal configurations and command sequences.
Real-robot experiments use the AGIBOT A3 humanoid robot.

\smallskip
\noindent\textbf{Baselines.}\enspace
SONIC~\cite{luo2026sonic} and UFO~\cite{ufo2026} use released G1 policies; UFO uses the Forward--Backward implementation (UFO-FB).
BeJEPA~\cite{yan2026bejepa} uses our reproduced actor without online model predictive control (MPC).
Our PLAT~\cite{wang2026plat} reproduction includes the Stage I PPO teacher and Stage II distilled policy, excluding Stage III latent residual reinforcement learning.
For sparse commands, SONIC's future-reference slots repeat the issued target.
Reference inputs are limited to the issued keyframes; intermediate frames are used for scoring.
For fixed spatial goals, SONIC and UFO retain their native interfaces without target XY input.

\smallskip
\noindent\textbf{Evaluation metrics.}\enspace
Tracking errors compare each control step with the reference motion at the same time.

\noindent\emph{1. Tracking accuracy.}\enspace
We report mean per-joint position error (\textbf{MPJPE}, cm) locally after removing pelvis translation while retaining orientation differences, and globally in world coordinates.
\textbf{Root XY error} (m) and \textbf{yaw error} (degrees) measure planar root-position distance and the shortest heading difference.
Terminal errors average these four quantities over the final 5\,s of fixed-goal trials.
For command-horizon analysis, \textbf{root-pose error} combines translation and heading:
\begin{equation}
e_{\mathrm{pose}}(t)=\sqrt{e_{xy}(t)^2+4r_{\mathrm{G1}}^2\sin^2\!\left(\tfrac{e_\psi(t)}{2}\right)}.
\label{eq:q1_pose_error}
\end{equation}
Here $r_{\mathrm{G1}}=0.1376$\,m, $e_{xy}$ is in meters and $e_\psi$ in radians; results are reported in cm.
Root-pose comparisons use the same heading-valid reference frames for all methods.

\noindent\emph{2. Goal acquisition and timing.}\enspace
\textbf{Arrival rate (AR)} is the fraction of complete commands jointly attaining local/global MPJPE $\leq20/50$\,cm, XY error $\leq0.5$\,m and yaw error $\leq45^\circ$ between issuance and the scheduled time plus 0.3\,s.
\textbf{On-time rate (OTR)} uses the same commands and thresholds, requiring first attainment within $\pm0.3$\,s of the scheduled time.
\textbf{Goal-reaching rate (Reach)} is the fraction of fixed-goal trials jointly attaining local/global MPJPE $\leq20/40$\,cm, XY error $\leq20$\,cm and yaw error $\leq45^\circ$ within the rollout.

\noindent\emph{3. Target retention.}\enspace
\textbf{Hold}$_5$ is the fraction of all fixed-goal trials maintaining XY error $\leq20$\,cm continuously for 5\,s after the first Reach event.
\textbf{Maximum holding error} ($E_{\max}$, cm) is the median across all short-command trials of the maximum XY error in this same 5\,s window.

\noindent\emph{4. Motion smoothness.}\enspace
\textbf{Joint jerk} is the RMS of $d^3q/dt^3$ over joints and time, reported in $10^3$\,rad/s$^3$.

\begin{figure*}[t]
\centering
\includegraphics[width=\textwidth]{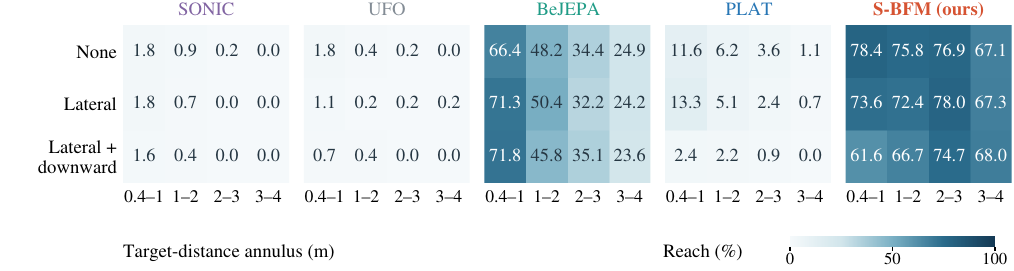}
\caption{\textbf{Goal-reaching rates across target distances and external perturbations.} Rows denote perturbation conditions and columns denote target-distance ranges. Cells report Reach (\%), with a shared 0--100\% color scale across methods.}
\label{fig:q2_arrival_heatmap}
\end{figure*}

\begin{table*}[!t]
\centering
\begin{minipage}[t]{\dimexpr(\textwidth-\columnsep)/2\relax}
\vspace{0pt}
\centering
\caption{\textbf{Fixed spatial goal reaching and retention without external perturbations.}}
\label{tab:q2_goalreach}
\footnotesize
\setlength{\tabcolsep}{1.5pt}
\renewcommand{\arraystretch}{1.08}
\begin{tabular*}{\linewidth}{@{\extracolsep{\fill}}lrrrrrr@{}}
\toprule
& \multicolumn{2}{c}{Attainment (\%)} & \multicolumn{4}{c}{Terminal error $\downarrow$} \\
\cmidrule(lr){2-3}\cmidrule(lr){4-7}
Method & Reach $\uparrow$ & Hold$_5$ $\uparrow$ & \shortstack{XY\\(m)} & \shortstack{Yaw\\($^\circ$)} & \shortstack{Local\\(cm)} & \shortstack{Global\\(cm)} \\
\midrule
SONIC~\cite{luo2026sonic} & 0.72 & 0.28 & 2.195 & 26.25 & 17.35 & 223.97 \\
UFO~\cite{ufo2026} & 0.61 & 0.00 & 2.491 & 91.62 & 24.24 & 250.68 \\
BeJEPA~\cite{yan2026bejepa} & 43.50 & 7.39 & 1.667 & 25.97 & 12.68 & 168.24 \\
PLAT~\cite{wang2026plat} & 5.61 & 4.72 & 1.779 & 100.85 & 40.66 & 194.16 \\
\textcolor{SBFMAccent}{\textbf{S-BFM (ours)}} & \textbf{74.56} & \textbf{35.22} & \textbf{0.257} & \textbf{21.56} & \textbf{10.89} & \textbf{29.88} \\
\bottomrule
\end{tabular*}
\end{minipage}\hfill
\begin{minipage}[t]{\dimexpr(\textwidth-\columnsep)/2\relax}
\vspace{0pt}
\centering
\caption{\textbf{P3 ablation: Goal retention under short remaining-time commands.}}
\label{tab:q3_p3_precision}
\footnotesize
\setlength{\tabcolsep}{3pt}
\renewcommand{\arraystretch}{1.08}
\begin{tabular*}{\linewidth}{@{\extracolsep{\fill}}lrrrr@{}}
\toprule
& \multicolumn{2}{c}{2\,m} & \multicolumn{2}{c}{3\,m} \\
\cmidrule(lr){2-3}\cmidrule(lr){4-5}
Training stages & \shortstack{Hold$_5$\\(\%)$\uparrow$} & \shortstack{$E_{\max}$\\(cm)$\downarrow$} & \shortstack{Hold$_5$\\(\%)$\uparrow$} & \shortstack{$E_{\max}$\\(cm)$\downarrow$} \\
\midrule
P1+P2 & 13.6 & 21.87 & 39.2 & 20.55 \\
\textcolor{SBFMAccent}{P1+P2+P3} & \textbf{92.0} & \textbf{19.81} & \textbf{89.6} & \textbf{19.73} \\
\bottomrule
\end{tabular*}
\end{minipage}
\end{table*}

\begin{figure}[t]
\centering
\includegraphics[width=0.90\columnwidth]{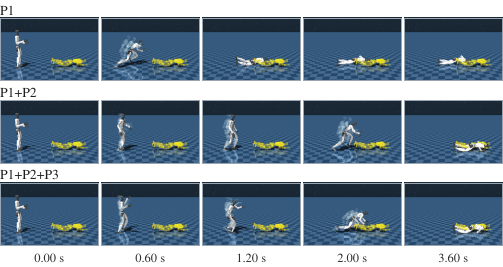}
\caption{\textbf{Qualitative comparison across training stages in simulation.} Transitions from a shared start to a prone goal 2\,m away. Gold: target; faint blue: past poses.}
\label{fig:q3_stage_qualitative}
\vspace{-5pt}
\end{figure}
\vspace{-5pt}

\subsection{Sparse Timed Keyframe Tracking}
\label{sec:q1}
To answer \textbf{Q1}, we evaluate tracking under successive timed keyframes using two command schedules.
\emph{Fixed} comprises 20 distinct horizons across four ranges: \emph{Short} spans 0.1--0.5\,s in 0.1\,s increments; \emph{Medium}, \emph{Long} and \emph{Extra-long} span 0.5--1.5\,s, 1.5--2.5\,s and 2.5--3.5\,s in 0.2\,s increments.
Shared boundaries are counted once.
\emph{Mix} samples $H=0.02K$ with $K$ uniform over $\{1,\ldots,125\}$.
Targets switch at their scheduled times.

\noindent\textbf{Coordinated posture and spatial tracking.}\enspace
Table~\ref{tab:q1_sparse} shows that S-BFM achieves the best tracking metrics under both Fixed and Mix, with global MPJPE of 38.35 and 37.60\,cm, respectively, and similar performance across schedules.
SONIC and UFO exhibit relatively small local errors but large world-space deviations.
BeJEPA and PLAT provide stronger baselines, yet S-BFM improves both posture and world-space tracking.

\noindent\textbf{Robustness to sparse guidance.}\enspace
Figure~\ref{fig:q1_horizon} shows that PLAT has low errors at short horizons, but its global body and root-pose errors rise sharply as reference updates become sparser.
S-BFM maintains flatter curves and the lowest errors across Extra-long horizons (2.5--3.5\,s), indicating more reliable world-space tracking under sparse guidance.

\subsection{Spatial Goal Reaching and Retention}
\label{sec:q2}
For \textbf{Q2}, we evaluate autonomous reaching of fixed, zero-velocity goals over 25\,s rollouts.
We pair 15 initial and 15 target configurations covering prone and seated poses, crouching, bowing and single-leg stance.
For each pair, two area-uniform target placements in each of four distance annuli (0.4--1, 1--2, 2--3 and 3--4\,m) yield 1,800 trials per method and perturbation condition, with 450 trials per annulus.
S-BFM and PLAT receive a 4\,s countdown that remains zero after expiration.
We evaluate unperturbed reaching and goal completion with torso forces applied at 1\,s for 0.2\,s.
Disturbances use a lateral impulse per unit mass of 0.5\,m/s, or stronger lateral/downward impulses of $1.5/-1.0$\,m/s.

\noindent\textbf{Goal acquisition and retention.}\enspace
S-BFM achieves the highest Reach and Hold$_5$ and the lowest terminal errors (Table~\ref{tab:q2_goalreach}).
Compared with BeJEPA, Reach increases from 43.5\% to 74.6\% and Hold$_5$ from 7.4\% to 35.2\%, while terminal XY error decreases from 1.667 to 0.257\,m.
The larger relative reduction in global than local MPJPE indicates more accurate world-space placement alongside posture matching.

\noindent\textbf{Long-range reaching under perturbations.}\enspace
Figure~\ref{fig:q2_arrival_heatmap} shows that S-BFM maintains high goal-completion rates over longer distances and under disturbances.
In the 3--4\,m annulus, Reach is 67.1\% for S-BFM versus 24.9\% for BeJEPA without perturbations, and 68.0\% versus 23.6\% under stronger lateral/downward perturbations.
These results show that S-BFM coordinates whole-body posture and world-space displacement to autonomously reach goals across distances and perturbation conditions.
\par
\endgroup

\begin{table*}[!t]
\centering
\caption{\textbf{Ablation of training stages for motion tracking and spatial goal control.} P1, P2 and P3 denote spatial grounding, time-conditioned distillation and generative fine-tuning. Time-conditioned indicates support for a specified arrival time.}
\label{tab:q3_stages}
\footnotesize
\setlength{\tabcolsep}{3pt}
\renewcommand{\arraystretch}{1.10}
\begin{tabular*}{\textwidth}{@{\extracolsep{\fill}}ccccrrrrrrrr@{}}
\toprule
\multicolumn{3}{c}{Training stages} & Capability & Dense & \multicolumn{5}{c}{Sparse timed tracking} & \multicolumn{2}{c}{Fixed-goal error} \\
\cmidrule(lr){1-3}\cmidrule(lr){4-4}\cmidrule(lr){5-5}\cmidrule(lr){6-10}\cmidrule(lr){11-12}
P1 & P2 & P3 & \shortstack{Time-\\conditioned} & \shortstack{Global\\(cm)$\downarrow$} & \shortstack{Local\\(cm)$\downarrow$} & \shortstack{Global\\(cm)$\downarrow$} & \shortstack{AR\\(\%)$\uparrow$} & \shortstack{OTR\\(\%)$\uparrow$} & \shortstack{Jerk\\($10^3$\,rad/s$^3$)$\downarrow$} & \shortstack{XY\\(m)$\downarrow$} & \shortstack{Yaw\\($^\circ$)$\downarrow$} \\
\midrule
$\checkmark$ & -- & -- & $\times$ & \textbf{34.65} & 9.81 & \textbf{36.62} & \textbf{90.63} & 4.63 & 1.38 & 0.739 & \textbf{8.75} \\
-- & $\checkmark$ & -- & $\checkmark$ & 130.39 & 14.64 & 188.10 & 31.16 & 2.13 & 1.36 & 3.502 & 80.87 \\
$\checkmark$ & $\checkmark$ & -- & $\checkmark$ & 37.00 & 9.34 & 38.95 & 84.67 & \textbf{17.35} & 1.24 & \textbf{0.597} & 8.99 \\
\midrule
$\checkmark$ & $\checkmark$ & $\checkmark$ & $\checkmark$ & 36.84 & \textbf{9.05} & 38.20 & 85.51 & 17.33 & \textbf{1.14} & 0.621 & 9.33 \\
\bottomrule
\end{tabular*}
\end{table*}

\begin{figure*}[!t]
  \centering
  \includegraphics[width=0.90\textwidth]{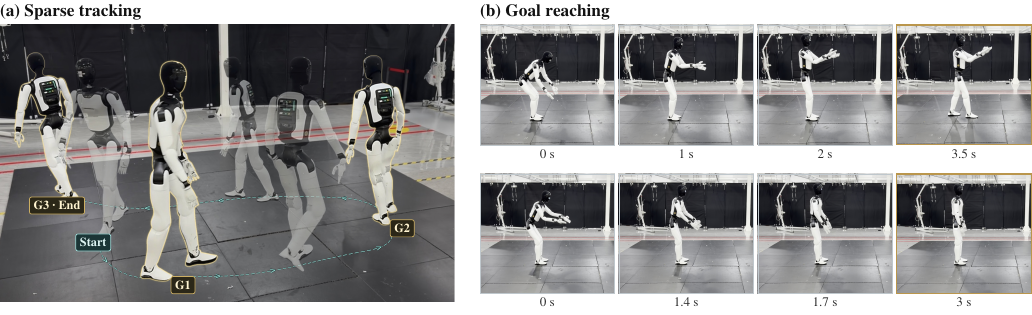}
  \caption{\textbf{Real-robot executions.} (a) Sparse tracking with a 2\,s command horizon; selected poses are highlighted in gold and intermediate poses are shown translucently. (b) Two continuous goal-reaching transitions, with elapsed video time shown below each frame.}
  \label{fig:real_robot}
\end{figure*}

\begin{figure*}[!t]
  \centering
  \includegraphics[width=0.90\textwidth]{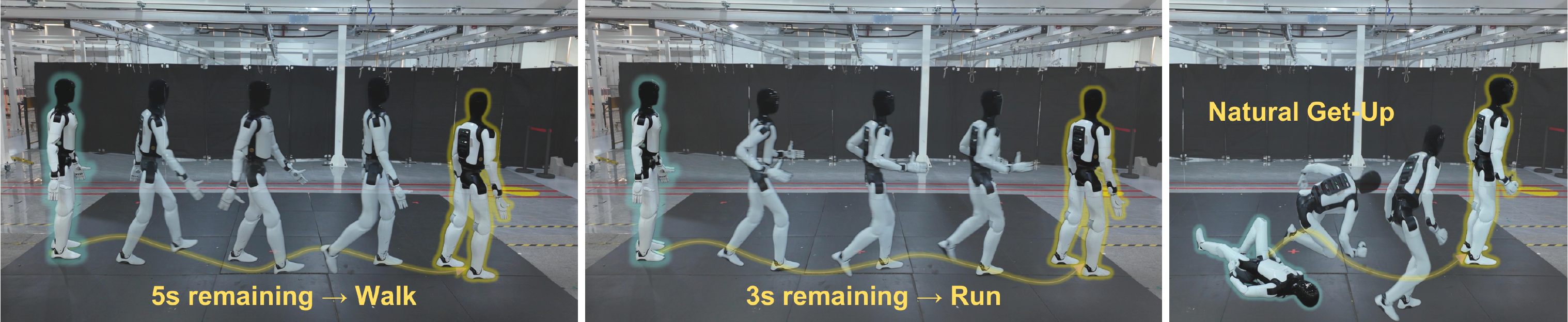}
  \caption{\textbf{Time-conditioned real-robot motions.}
  Left and middle: the robot walks with 5\,s remaining and runs with 3\,s remaining to reach the target.
  Right: the robot autonomously organizes a coordinated whole-body transition from lying on the ground to standing.}
  \label{fig:real_robot_temporal}
\end{figure*}

\subsection{Ablation Study}
\label{sec:q3}

To answer \textbf{Q3}, we compare P1, P2 without spatial grounding, P1+P2, and P1+P2+P3 on the same motion tracking and fixed-goal tasks (Table~\ref{tab:q3_stages}). P2 alone uses the released UFO policy as its base.

\noindent\textbf{Contributions of the training stages.}\enspace
P1 alone provides effective motion and world-space position tracking, but its on-time rate is only 4.63\%. Without explicit supervision of transition motions and execution timing, it may adopt the prone goal posture while still far from the target location (\cref{fig:q3_stage_qualitative}). This premature posture change can restrict subsequent locomotion, making it difficult to complete the remaining displacement and potentially preventing the robot from reaching the target.
P2 without P1 yields a terminal XY error of 3.502\,m, indicating that reference-motion distillation alone has not provided this variant with reliable goal-position control in world coordinates. Combining P1 and P2 raises the on-time rate to 17.35\% and reduces terminal XY error to 0.597\,m; the robot also approaches the target before lying prone. These results show that time-conditioned distillation built on the spatial behavior model coordinates target-position control with transition organization. Adding P3 further reduces sparse tracking errors and joint jerk.

\noindent\textbf{Target retention under short commands.}\enspace
To assess P3 under short remaining-time commands, we fine-tune the same P1+P2 checkpoint for 10M steps using equal proportions of real and generated references. We evaluate standing goals at 2 and 3\,m with remaining-time commands from 1 to 3\,s in 0.5\,s increments, using 250 paired trials per distance. Table~\ref{tab:q3_p3_precision} shows that P3 increases Hold$_5$ from 13.6\% to 92.0\% at 2\,m and from 39.2\% to 89.6\% at 3\,m, while reducing $E_{\max}$ at both distances. 
P3 thus substantially improves the reliability of target-position retention after arrival under short remaining-time commands.\par

\subsection{Real-Robot Experiments}
\label{sec:real_robot}
To answer \textbf{Q4}, we deploy the complete three-stage S-BFM on the full-size AGIBOT A3 humanoid.
Figure~\ref{fig:real_robot} presents representative sparse tracking and transitions between whole-body goal postures. Figure~\ref{fig:real_robot_temporal} illustrates time-adaptive locomotion and natural get-up.
From the issued goals, the robot organizes intermediate steps and coordinates adjustments of its arms, legs and trunk throughout the transitions.
Together with the goal-posture examples in Fig.~\ref{fig:sbfm_teaser}, these demonstrations show that the sparse-goal interface supports coordinated whole-body control on physical hardware.

\section{Conclusion, Limitation and Future Work}
\label{sec:conclusion}

S-BFM enables humanoid robots to organize whole-body motion and execution pace from sparse spatiotemporal goals, without dense reference trajectories at deployment. Spatial grounding and time-conditioned distillation support accurate spatial control and coordinated transitions. Generative fine-tuning extends control across a broader range of arrival-time commands. Our current approach has three main limitations.

First, these capabilities are acquired through three successive training stages, which complicates the training pipeline. Future work will explore incorporating remaining time and explicit terminal conditions into forward–backward representations and Bellman updates. Combining this formulation with goal- and time-conditioned adversarial motion priors may enable spatial control, natural transitions, and timing control to be learned within a single stage.

Second, integration with vision–language–action and world-action models remains unexplored. We will investigate interfaces that allow these models to specify executable spatial goals and arrival times. This will involve developing suitable real-robot data collection procedures and adapting high-level outputs and training objectives.

Third, we have not studied how high-level architectures should be redesigned around S-BFM’s control capabilities. Future work will explore architectures that focus on scene understanding, task reasoning, and behavior goals critical to task success, while S-BFM organizes intermediate motion. Reachability and execution feedback could further help high-level models refine their goals and timing decisions.
\FloatBarrier
\pagebreak[4]

{
    \small
    \setlength{\bibsep}{1pt}
    \bibliographystyle{ieeenat_fullname}
    \bibliography{main}

@inproceedings{li2026bfmzero,
  title = {{BFM-Zero: A Promptable Behavioral Foundation Model for Humanoid Control Using Unsupervised Reinforcement Learning}},
  author = {Yitang Li and Zhengyi Luo and Tonghe Zhang and Cunxi Dai and Anssi Kanervisto and Andrea Tirinzoni and Haoyang Weng and Kris Kitani and Mateusz Guzek and Ahmed Touati and Alessandro Lazaric and Matteo Pirotta and Guanya Shi},
  year = {2026},
  booktitle = {International Conference on Learning Representations},
  url = {https://arxiv.org/abs/2511.04131},
}

@article{yan2026bejepa,
  title = {{BeJEPA: A Goal-Conditioned Humanoid Behavior World Model via Unsupervised RL}},
  author = {Liyun Yan and Haoyang Li and Yang Zhang and Jialing Zhu and Zhanxiang Cao and Zihao Zhou and Keqi Zhu and Yizhi Chen and Huang Chuxuan and Shengcheng Fu and Zhong Jiangwei and Qiao Sun and Chen Zhang and Yue Gao},
  year = {2026},
  journal = {arXiv preprint arXiv:2610.23127},
  url = {https://arxiv.org/abs/2610.23127},
  eprint = {2610.23127},
  archivePrefix = {arXiv},
}

@article{peng2018deepmimic, eid = {143},
  title   = {{DeepMimic}: Example-Guided Deep Reinforcement Learning of Physics-Based Character Skills},
  author  = {Peng, Xue Bin and Abbeel, Pieter and Levine, Sergey and van de Panne, Michiel},
  journal = {ACM Transactions on Graphics},
  volume  = {37},
  number  = {4},
  year    = {2018}
}

@inproceedings{luo2023phc,
  title     = {Perpetual Humanoid Control for Real-time Simulated Avatars},
  author    = {Luo, Zhengyi and Cao, Jinkun and Winkler, Alexander and Kitani, Kris and Xu, Weipeng},
  booktitle = {IEEE/CVF International Conference on Computer Vision (ICCV)},
  year      = {2023}
}

@inproceedings{he2024omnih2o,
  title     = {{OmniH2O}: Universal and Dexterous Human-to-Humanoid Whole-Body Teleoperation and Learning},
  author    = {He, Tairan and Luo, Zhengyi and He, Xialin and Xiao, Wenli and Zhang, Chong and Zhang, Weinan and Kitani, Kris and Liu, Changliu and Shi, Guanya},
  booktitle = {Conference on Robot Learning (CoRL)},
  year      = {2024}
}

@inproceedings{cheng2024exbody,
  title     = {Expressive Whole-Body Control for Humanoid Robots},
  author    = {Cheng, Xuxin and Ji, Yandong and Chen, Junming and Yang, Ruihan and Yang, Ge and Wang, Xiaolong},
  booktitle = {Robotics: Science and Systems (RSS)},
  year      = {2024}
}

@inproceedings{schaul2015uvfa,
  title     = {Universal Value Function Approximators},
  author    = {Schaul, Tom and Horgan, Daniel and Gregor, Karol and Silver, David},
  booktitle = {International Conference on Machine Learning (ICML)},
  year      = {2015}
}

@inproceedings{touati2023zeroshot,
  title     = {Does Zero-Shot Reinforcement Learning Exist?},
  author    = {Touati, Ahmed and Rapin, J{\'e}r{\'e}my and Ollivier, Yann},
  booktitle = {International Conference on Learning Representations (ICLR)},
  year      = {2023}
}

@inproceedings{tevet2023mdm,
  title     = {Human Motion Diffusion Model},
  author    = {Tevet, Guy and Raab, Sigal and Gordon, Brian and Shafir, Yonatan and Cohen-Or, Daniel and Bermano, Amit H.},
  booktitle = {International Conference on Learning Representations (ICLR)},
  year      = {2023}
}

@inproceedings{cohan2024condmdi,
  title     = {Flexible Motion In-betweening with Diffusion Models},
  author    = {Cohan, Setareh and Tevet, Guy and Reda, Daniele and Peng, Xue Bin and van de Panne, Michiel},
  booktitle = {ACM SIGGRAPH Conference Papers},
  year      = {2024}
}

@article{rupf2026bfmtrack,
  title = {{BFMTrack: Latent Sequence Optimization for Physics-Based Motion Tracking with Behavioral Foundation Models}},
  author = {Thomas Rupf and Agon Serifi and David M{\"u}ller and Sammy Christen and Ruben Grandia and Espen Knoop and Moritz B{\"a}cher},
  year = {2026},
  journal = {arXiv preprint arXiv:2606.25056},
  url = {https://arxiv.org/abs/2606.25056},
  eprint = {2606.25056},
  archivePrefix = {arXiv},
}

@article{wang2026plat,
  title = {{PLAT: Sparse Timed Keyframe Motion Tracking for Humanoid Control via Privileged Latent Transition Learning}},
  author = {Wang, Zepeng and Wang, Jiangxing and Ma, Chao and Shi, Xiaochuan and Lu, Zongqing},
  year = {2026},
  journal = {arXiv preprint arXiv:2609.25754},
  url = {https://arxiv.org/abs/2609.25754},
  eprint = {2609.25754},
  archivePrefix = {arXiv},
}

@article{chen2026kino,
  title = {{KINO: A Keyframe Interface for VLM Planning and Whole-Body Control in Humanoid Loco-Manipulation}},
  author = {Sitong Chen and Fatemeh Zargarbashi and Jin Cheng and Tianxu An and Stelian Coros},
  year = {2026},
  journal = {arXiv preprint arXiv:2609.18869},
  url = {https://arxiv.org/abs/2609.18869},
  eprint = {2609.18869},
  archivePrefix = {arXiv},
}

@article{tessler2024maskedmimic, volume = {43}, number = {6}, eid = {209}, pages = {1--21},
  title = {{MaskedMimic: Unified Physics-Based Character Control Through Masked Motion Inpainting}},
  author = {Chen Tessler and Yunrong Guo and Ofir Nabati and Gal Chechik and Xue Bin Peng},
  year = {2024},
  journal = {ACM Transactions on Graphics},
  url = {https://arxiv.org/abs/2409.14393},
}

@article{harvey2020inbetweening, eid = {60},
  title = {{Robust Motion In-betweening}},
  author = {F{\'e}lix G. Harvey and Mike Yurick and Derek Nowrouzezahrai and Christopher Pal},
  year = {2020},
  journal = {ACM Transactions on Graphics},
  volume = {39},
  number = {4},
  doi = {10.1145/3386569.3392480},
  url = {https://arxiv.org/abs/2102.04942},
}

@inproceedings{pong2018tdm,
  title = {{Temporal Difference Models: Model-Free Deep RL for Model-Based Control}},
  author = {Vitchyr Pong and Shixiang Gu and Murtaza Dalal and Sergey Levine},
  year = {2018},
  booktitle = {International Conference on Learning Representations},
  url = {https://arxiv.org/abs/1802.09081},
}

@article{luo2026sonic,
  title = {{SONIC: Supersizing Motion Tracking for Natural Humanoid Whole-Body Control}},
  author = {Zhengyi Luo and Ye Yuan and Tingwu Wang and Chenran Li and Fernando Casta{\~n}eda and Sirui Chen and Zi-Ang Cao and Jiefeng Li and David Minor and Qingwei Ben and Jinhyung Park and David Sami and Zi Wang and Xingye Da and Runyu Ding and Cyrus Hogg and Lina Song and Edy Lim and Eugene Jeong and Tairan He and Haoru Xue and Wenli Xiao and Simon Yuen and Jan Kautz and Yan Chang and Umar Iqbal and Linxi "Jim" Fan and Yuke Zhu},
  year = {2026},
  journal = {Science Robotics},
  volume = {11},
  number = {117},
  pages = {eaed4592},
  doi = {10.1126/scirobotics.aed4592},
  url = {https://arxiv.org/abs/2511.07820},
}

@article{li2026pgmt,
  title = {{PGMT: Perceptive General Motion Tracking for Humanoid Robots}},
  author = {Hongyi Li and Li Peizhuo and Yucheng Tao and Ze Wang and Fangzhou Xu and Jinyi Chen and Yanyan Yuan and Dapeng Jia and Yongbin Jin and Mingfeng Fan and Guillaume Sartoretti and Hongtao Wang},
  year = {2026},
  journal = {arXiv preprint arXiv:2609.08511},
  url = {https://arxiv.org/abs/2609.08511},
  eprint = {2609.08511},
  archivePrefix = {arXiv},
}

@article{ma2026extremergmt,
  title = {{Extreme-RGMT: Continual Learning of Highly Dynamic Skills for Robust Generalist Humanoid Control}},
  author = {Yubiao Ma and Han Yu and Kai Guo and Changtai Lv and Zhengquan Mao and Boyang Xing and Xuemei Ren and Dongdong Zheng},
  year = {2026},
  journal = {arXiv preprint arXiv:2607.20110},
  url = {https://arxiv.org/abs/2607.20110},
  eprint = {2607.20110},
  archivePrefix = {arXiv},
}

@article{xu2026interprior,
  title = {{InterPrior: Scaling Generative Control for Physics-Based Human-Object Interactions}},
  author = {Sirui Xu and Samuel Schulter and Morteza Ziyadi and Xialin He and Xiaohan Fei and Yu-Xiong Wang and Liangyan Gui},
  year = {2026},
  journal = {arXiv preprint arXiv:2602.06035},
  url = {https://arxiv.org/abs/2602.06035},
  eprint = {2602.06035},
  archivePrefix = {arXiv},
}

@inproceedings{liao2026supermimic,
  title = {{SuperMimic: Scaling Humanoid Whole-Body Control with One Differentiable World Model}},
  author = {Qiayuan Liao and Yiyang Shao and Junfeng Long and Kevin Zakka and Roberto Horowitz and Pieter Abbeel and Koushil Sreenath},
  year = {2026},
  booktitle = {Conference on Robot Learning},
  url = {https://arxiv.org/abs/2610.28480},
}

@article{ye2026predactor,
  title = {{PredActor: Predictive Action Diffusion for Steerable Onboard Humanoid Control}},
  author = {Lei Ye and Haibo Gao and Yitang Li and Peng Xu and Zetong Jing and Junhan Sun and Fanrong Dong and Ziqi Han and Xue Wang and Jianhua Sun and Cewu Lu and Hao Zhao and Liang Ding},
  year = {2026},
  journal = {arXiv preprint arXiv:2609.24840},
  url = {https://arxiv.org/abs/2609.24840},
  eprint = {2609.24840},
  archivePrefix = {arXiv},
}

@article{xie2026textop,
  title = {{TextOp: Real-time Interactive Text-Driven Humanoid Robot Motion Generation and Control}},
  author = {Weiji Xie and Jiakun Zheng and Jinrui Han and Jiyuan Shi and Weinan Zhang and Chenjia Bai and Xuelong Li},
  year = {2026},
  journal = {arXiv preprint arXiv:2602.07439},
  url = {https://arxiv.org/abs/2602.07439},
  eprint = {2602.07439},
  archivePrefix = {arXiv},
}

@article{zhao2026ardy,
  title = {{ARDY: Autoregressive Diffusion with Hybrid Representation for Interactive Human Motion Generation}},
  author = {Kaifeng Zhao and Mathis Petrovich and Haotian Zhang and Tingwu Wang and Siyu Tang and Davis Rempe},
  year = {2026},
  journal = {ACM Transactions on Graphics},
  volume = {45},
  number = {4},
  eid = {86},
  doi = {10.1145/3811284},
  url = {https://arxiv.org/abs/2607.08741},
}

@article{rempe2026kimodo,
  title = {{Kimodo: Scaling Controllable Human Motion Generation}},
  author = {Davis Rempe and Mathis Petrovich and Ye Yuan and Haotian Zhang and Xue Bin Peng and Yifeng Jiang and Tingwu Wang and Umar Iqbal and David Minor and Michael de Ruyter and Jiefeng Li and Chen Tessler and Edy Lim and Eugene Jeong and Sam Wu and Ehsan Hassani and Michael Huang and Jin-Bey Yu and Chaeyeon Chung and Lina Song and Olivier Dionne and Jan Kautz and Simon Yuen and Sanja Fidler},
  year = {2026},
  journal = {arXiv preprint arXiv:2603.15546},
  url = {https://arxiv.org/abs/2603.15546},
  eprint = {2603.15546},
  archivePrefix = {arXiv},
}

@inproceedings{xie2026grail,
  title = {{GRAIL: Generating Humanoid Loco-Manipulation from World Models}},
  author = {Tianyi Xie and Haotian Zhang and Jinhyung Park and Zi Wang and Bowen Wen and Jiefeng Li and Xueting Li and Qingwei Ben and Haoyang Weng and Yufei Ye and David Minor and Tingwu Wang and Chenfanfu Jiang and Sanja Fidler and Jan Kautz and Linxi Fan and Yuke Zhu and Zhengyi Luo and Umar Iqbal and Ye Yuan},
  year = {2026},
  booktitle = {Conference on Robot Learning},
  url = {https://arxiv.org/abs/2606.05160},
}

@article{ze2025twist2,
  title = {{TWIST2: Scalable, Portable, and Holistic Humanoid Data Collection System}},
  author = {Yanjie Ze and Siheng Zhao and Weizhuo Wang and Angjoo Kanazawa and Rocky Duan and Pieter Abbeel and Guanya Shi and Jiajun Wu and C. Karen Liu},
  year = {2025},
  journal = {arXiv preprint arXiv:2511.02832},
  url = {https://arxiv.org/abs/2511.02832},
  eprint = {2511.02832},
  archivePrefix = {arXiv},
}

@inproceedings{ross2011dagger,
  title = {{A Reduction of Imitation Learning and Structured Prediction to No-Regret Online Learning}},
  author = {Ross, Stephane and Gordon, Geoffrey and Bagnell, Drew},
  booktitle = {Proceedings of the Fourteenth International Conference on Artificial Intelligence and Statistics},
  pages = {627--635},
  year = {2011},
  volume = {15},
  series = {Proceedings of Machine Learning Research},
  publisher = {PMLR},
  url = {https://proceedings.mlr.press/v15/ross11a.html},
}

@inproceedings{touati2021fb,
 author = {Touati, Ahmed and Ollivier, Yann},
 booktitle = {Advances in Neural Information Processing Systems},
 editor = {M. Ranzato and A. Beygelzimer and Y. Dauphin and P.S. Liang and J. Wortman Vaughan},
 pages = {13--23},
 publisher = {Curran Associates, Inc.},
 title = {Learning One Representation to Optimize All Rewards},
 url = {https://proceedings.neurips.cc/paper_files/paper/2021/file/003dd617c12d444ff9c80f717c3fa982-Paper.pdf},
 volume = {34},
 year = {2021}
}

@article{clark2015automaticity,
  author = {Clark, David J.},
  title = {Automaticity of walking: functional significance, mechanisms, measurement and rehabilitation strategies},
  journal = {Frontiers in Human Neuroscience},
  volume = {9},
  pages = {246},
  year = {2015},
  doi = {10.3389/fnhum.2015.00246},
  url = {https://www.frontiersin.org/journals/human-neuroscience/articles/10.3389/fnhum.2015.00246/full},
}

@inproceedings{he2024learning,
  title = {{Learning Human-to-Humanoid Real-Time Whole-Body Teleoperation}},
  author = {Tairan He and Zhengyi Luo and Wenli Xiao and Chong Zhang and Kris Kitani and Changliu Liu and Guanya Shi},
  booktitle = {IEEE/RSJ International Conference on Intelligent Robots and Systems},
  year = {2024},
  url = {https://arxiv.org/abs/2403.04436},
}

@article{ji2024exbody2,
  title = {{ExBody2: Advanced Expressive Humanoid Whole-Body Control}},
  author = {Mazeyu Ji and Xuanbin Peng and Fangchen Liu and Jialong Li and Ge Yang and Xuxin Cheng and Xiaolong Wang},
  journal = {arXiv preprint arXiv:2412.13196},
  year = {2024},
  url = {https://arxiv.org/abs/2412.13196},
}

@misc{ufo2026,
  author       = {{RoboParty Lab Team}},
  title        = {{UFO: An Unsupervised Reinforcement Learning Framework for Humanoid Control}},
  year         = {2026},
  howpublished = {\url{https://github.com/Roboparty/UFO}},
  note         = {Project page: \url{https://roboparty.github.io/UFO/}}
}

@misc{chen2026physicsguidedbiomechanicalgaitadaptation,
      title={Physics-Guided Biomechanical Gait Adaptation for Humanoid Locomotion on Extreme Sloped Terrains}, 
      author={Xuanyu Chen and Mohan Liu and Dengchen Mei and Zhihao Gu and Haitian Zhang and Kaimin Mao and Haiyue Zhu and Shijun Yan and Lin Wang},
      year={2026},
      eprint={2607.07830},
      archivePrefix={arXiv},
      primaryClass={cs.RO},
      url={https://arxiv.org/abs/2607.07830}, 
}

@misc{hu2026maskillblenderdecentralizedwholebodycoordination,
      title={{MASkillBlender}: Decentralized Whole-Body Coordination for Multi-Humanoid Loco-Manipulation via Skill Blending}, 
      author={Yifan Hu and Luhang Hong and Mingkang Long and Danning Wang and Chengfeng Jia and Rong Su and Junjie Fu and Guanghui Wen},
      year={2026},
      eprint={2610.01102},
      archivePrefix={arXiv},
      primaryClass={cs.RO},
      url={https://arxiv.org/abs/2610.01102}, 
}

@article{tan2026slmp,
  title = {Spherical Latent Motion Prior for Physics-Based Simulated Humanoid Control},
  author = {Jing Tan and Weisheng Xu and Xiangrui Jiang and Jiaxi Zhang and Kun Yang and Kai Wu and Jiaqi Xiong and Shiting Chen and Yangfan Li and Yixiao Feng and Yuetong Fang and Yujia Zou and Yiqun Song and Renjing Xu},
  journal = {arXiv preprint arXiv:2603.01294},
  year = {2026},
  url = {https://arxiv.org/abs/2603.01294},
}

@article{tan2026farm,
  title = {{FARM}: Frame-Accelerated Augmentation and Residual Mixture-of-Experts for Physics-Based High-Dynamic Humanoid Control},
  author = {Jing Tan and Shiting Chen and Yangfan Li and Weisheng Xu and Renjing Xu},
  journal = {Proceedings of the AAAI Conference on Artificial Intelligence},
  volume = {40},
  number = {22},
  pages = {18575--18583},
  year = {2026},
  doi = {10.1609/aaai.v40i22.38924},
  url = {https://ojs.aaai.org/index.php/AAAI/article/view/38924},
}

@article{yang2026uniquenft,
  title = {{UniqueNFT}: Uniqueness Protection of Digital Assets in Decentralized Web},
  author = {Kun Yang and Haihan Duan and Yubo Zhao and Jiale Cheng and Runhao Zeng and Xiping Hu},
  journal = {ACM Transactions on the Web},
  volume = {20},
  number = {1},
  pages = {1--30},
  year = {2026},
  doi = {10.1145/3779414},
  url = {https://doi.org/10.1145/3779414},
}

@misc{han2026ibfm,
  title = {{I-BFM}: Reward-Conditioned Robust Humanoid Interaction via Unsupervised Reinforcement Learning},
  author = {Ziqi Han and Yitang Li and Junhan Sun and Fanrong Dong and Yaojie Shen and Lei Ye and Zetong Jing and Yongqi Zhang and Yiming Zhang and Xue Wang and Hao Zhao},
  year = {2026},
  eprint = {2610.06129},
  archivePrefix = {arXiv},
  primaryClass = {cs.RO},
  url = {https://arxiv.org/abs/2610.06129},
}

@misc{wang2026pbfm,
  title = {{P-BFM}: Perceptive Behavioral Foundation Models for Humanoid Control via Unsupervised Reinforcement Learning},
  author = {Xue Wang and Yitang Li and Yiming Zhang and Siqin Wang and Junhan Sun and Zhihao Liu and Yaojie Shen and Jingbo Yuan and Hao Zhao and Qingdu Li},
  year = {2026},
  eprint = {2610.15833},
  archivePrefix = {arXiv},
  primaryClass = {cs.RO},
  url = {https://arxiv.org/abs/2610.15833},
}

@misc{chen2025gmtgeneralmotiontracking,
      title={{GMT}: General Motion Tracking for Humanoid Whole-Body Control}, 
      author={Zixuan Chen and Mazeyu Ji and Xuxin Cheng and Xuanbin Peng and Xue Bin Peng and Xiaolong Wang},
      year={2025},
      eprint={2506.14770},
      archivePrefix={arXiv},
      primaryClass={cs.RO},
      url={https://arxiv.org/abs/2506.14770}, 
}

@misc{ze2025twistteleoperatedwholebodyimitation,
      title={{TWIST}: Teleoperated Whole-Body Imitation System}, 
      author={Yanjie Ze and Zixuan Chen and João Pedro Araújo and Zi-ang Cao and Xue Bin Peng and Jiajun Wu and C. Karen Liu},
      year={2025},
      eprint={2505.02833},
      archivePrefix={arXiv},
      primaryClass={cs.RO},
      url={https://arxiv.org/abs/2505.02833}, 
}

@misc{chen2026blinkmimictraininghumanoidmotion,
      title={{BlinkMimic}: Training a Humanoid Motion Tracker in Minutes on a Single {GPU}}, 
      author={Fuyi Chen},
      year={2026},
      eprint={2611.16422},
      archivePrefix={arXiv},
      primaryClass={cs.RO},
      url={https://arxiv.org/abs/2611.16422}, 
}

@article{he2025asap,
          title={{ASAP}: Aligning Simulation and Real-World Physics for Learning Agile Humanoid Whole-Body Skills},
          author={He, Tairan and Gao, Jiawei and Xiao, Wenli and Zhang, Yuanhang and Wang, Zi and Wang, Jiashun and Luo, Zhengyi and He, Guanqi and Sobanbabu, Nikhil and Pan, Chaoyi and Yi, Zeji and Qu, Guannan and Kitani, Kris and Hodgins, Jessica and Fan, Linxi "Jim" and Zhu, Yuke and Liu, Changliu and Shi, Guanya},
          journal={arXiv preprint arXiv:2502.01143},
          year={2025}
        }

@misc{harithas2026braceadaptingwholebodyreferences,
      title={{BRACE}: Adapting Whole-Body References for Force and Terrain Aware Humanoid Motion Tracking}, 
      author={Sudarshan Harithas and Chen Yu and Juan Borbon and Shubhankar Mondal and Winston Zha and Srinath Sridhar and Dingqi Zhang and Jiuguang Wang},
      year={2026},
      eprint={2610.07052},
      archivePrefix={arXiv},
      primaryClass={cs.RO},
      url={https://arxiv.org/abs/2610.07052}, 
}

@misc{zargarbashi2024robotkeyframinglearninglocomotionhighlevel,
      title={{RobotKeyframing}: Learning Locomotion with High-Level Objectives via Mixture of Dense and Sparse Rewards}, 
      author={Fatemeh Zargarbashi and Jin Cheng and Dongho Kang and Robert Sumner and Stelian Coros},
      year={2024},
      eprint={2407.11562},
      archivePrefix={arXiv},
      primaryClass={cs.RO},
      url={https://arxiv.org/abs/2407.11562}, 
}

@misc{huang2025adaptablehumanoidcontroladaptive,
      title={Towards Adaptable Humanoid Control via Adaptive Motion Tracking}, 
      author={Tao Huang and Huayi Wang and Junli Ren and Kangning Yin and Zirui Wang and Xiao Chen and Feiyu Jia and Wentao Zhang and Junfeng Long and Jingbo Wang and Jiangmiao Pang},
      year={2025},
      eprint={2510.14454},
      archivePrefix={arXiv},
      primaryClass={cs.RO},
      url={https://arxiv.org/abs/2510.14454}, 
}
}


\end{document}